\PassOptionsToPackage{dvipsnames}{xcolor}
\documentclass{article}
\usepackage{arxiv}
\usepackage{natbib}[numbers,sort,compress]

\usepackage{wrapfig}

\usepackage{multirow,mathtools }

\usepackage{pifont}
\usepackage{color, colortbl}

\usepackage{blindtext}
\usepackage{lipsum}

\usepackage{multirow}
\usepackage{listings}

\usepackage{bbm}

\usepackage [english]{babel}
\usepackage[autostyle, english = american]{csquotes}

\usepackage{pifont}
\usepackage{url}
\usepackage[most]{tcolorbox}

\usepackage{lipsum}
\usepackage{soul}
\usepackage{xcolor}
\usepackage{wrapfig}
\usepackage{multirow,mathtools } 

\usepackage{adjustbox}
\MakeOuterQuote{"}

\usepackage{microtype}
\usepackage{graphicx}
\usepackage{subfigure}
\usepackage{booktabs} % for professional tables
\usepackage{hyperref}

\usepackage{amsmath}

\usepackage[capitalize,noabbrev]{cleveref}

\usepackage{nicefrac}       % compact symbols for 1/2, etc.

\usepackage{tablefootnote}

\usepackage{float} 
\usepackage{algorithm}
\usepackage{algpseudocode}

\newcommand{\Def}[0]{\mathrel{\mathop:}=}

\usepackage{pifont}
\newcommand{\cmark}{\textcolor{green}{\ding{51}}}
\newcommand{\xmark}{\textcolor{red}{\ding{55}}}

\usepackage{color, colortbl}
\definecolor{Gray}{gray}{0.93}
\definecolor{Orange}{rgb}{1,0.5,0}
\definecolor{DGray}{gray}{0.83}
\definecolor{LightCyan}{rgb}{0.88,1,1}

\definecolor{WarnREd}{rgb}{1,0.4,0.4}
\definecolor{WarnOrange}{rgb}{1,0.682,0.502}
\definecolor{WarnPink}{rgb}{0.9176, 0.7215, 0.7215}
\definecolor{GoodGreen}{rgb}{0.5019, 0.9215, 0.6039}

\usepackage[T1]{fontenc}

\definecolor{styleblue}{HTML}{504099}
\definecolor{mypurple}{HTML}{9391ff}

\definecolor{bluegray}{rgb}{0.4, 0.6, 0.8}
\definecolor{ceruleanblue}{rgb}{0.16, 0.32, 0.75}

\hypersetup{
colorlinks=true,
citecolor=ceruleanblue,
linkcolor=ceruleanblue,
urlcolor=black
}

\usepackage{xcolor}
\usepackage{float}      
\usepackage{caption}    

\lstdefinestyle{simple}{
  language=Python,
  basicstyle=\ttfamily\scriptsize, % 比 footnotesize 更小
  numbers=none,                  
  frame=single,
  rulecolor=\color{black},
  backgroundcolor=\color[gray]{0.96},
  showstringspaces=false,
  keepspaces=true,
  columns=fullflexible,
  breaklines=true,
  breakatwhitespace=true,
  tabsize=2
}

\usepackage{amsmath,amsfonts,bm}

\def\eqref#1{(\ref{#1})}
\def\1{\bm{1}}

\DeclareMathAlphabet{\mathsfit}{\encodingdefault}{\sfdefault}{m}{sl}
\SetMathAlphabet{\mathsfit}{bold}{\encodingdefault}{\sfdefault}{bx}{n}

\DeclareMathOperator*{\argmin}{arg\,min}

\DeclareMathOperator*{\minimize}{\text{minimize}}

\newcommand{\btheta}{{\boldsymbol{\theta}}}

\usepackage{times}
\usepackage{latexsym}

\usepackage[T1]{fontenc}

\usepackage{microtype}

\usepackage{inconsolata}

\usepackage{graphicx}
\usepackage{booktabs}
\usepackage{float} 
\usepackage{adjustbox}
\usepackage{multirow}
\usepackage{wrapfig} 
\usepackage{fontawesome5}

\everydisplay{\small}

\title{Train Locally, Learn Together: Distributed Multi-Agent Training via ADMM}

\author{%
Yiwei Chen$^{1}$\quad Changsheng Wang$^{1}$\quad Yicheng Lang$^{1}$\\
\textbf{Qing Qu$^{2}$\quad Alfred O. Hero$^{2}$\quad Sijia Liu$^{1}$}\\
$^{1}$Michigan State University \quad
$^{2}$University of Michigan, Ann Arbor
}
\date{\vspace{-6mm}%
\texttt{\href{https://github.com/OPTML-Group/Distributed-MAS}{\faGithub\ Code}}}

\begin{document}
\maketitle
%\vspace{-4mm}

\begin{abstract}
Multi-agent systems (MAS) can outperform individual language-model agents by combining specialized roles, diverse reasoning processes, and collaborative decision-making. 
% However, training and governing an MAS as a single centralized system requires one trainer to hold every agents’ data, parameters, and optimization states, incurring substantial resource overhead and creating a potential single point of failure. 
However, collaborative MAS training is typically centralized, requiring a common trainer to aggregate role-local trajectories and learning signals and maintain model and optimization states.
This motivates a {distributed multi-agent learning} paradigm in which agents improve system-level capabilities through role-local learning and decentralized communication, paving the way toward scalable, resilient, and self-organizing MAS.
To this end, we introduce, to the best of our knowledge, the first optimization-grounded framework for distributed MAS training based on consensus \textit{alternating direction method of multipliers} (ADMM).
 We formulate collaborative MAS learning as a consensus-constrained finite-sum optimization problem, naturally lending itself to distributed optimization via ADMM. 
The proposed approach alternates role-local learning with inter-agent consensus using distributed communication protocols such as ring all-reduce, enabling joint optimization while keeping each agent's data and training state local.
We evaluate our framework on distributed safety alignment and general reasoning tasks. 
% We show that ADMM-based MAS training achieves gains comparable to centralized joint training without requiring a centralized trainer. 
% Notably, distributed system-level alignment improves MAS safety even in the presence of a compromised agent. 
Under compromised-agent settings, ADMM can reduce HarmBench attack success rate (ASR) by up to 26.4\%, while achieving reasoning gains comparable to centralized training.
These results show that distributed MAS training retains the benefits of centralized joint optimization without requiring a centralized trainer.
% Code is available at \href{https://github.com/OPTML-Group/Distributed-MAS}{\textcolor{magenta}{\texttt{https://github.com/OPTML-Group/Distributed-MAS}}}.
\end{abstract}

\section{Introduction}
\label{sec:intro}

Large language models (LLMs) are increasingly organized into multi-agent systems (MAS), in which agents assume specialized \textit{roles} and communicate in natural language or latent messages to solve complex tasks collaboratively~\citep{li2023camel,wu2024autogen,pmlr-v235-du24e,zou2026latent}. 
By coordinating complementary roles, diverse reasoning processes, and iterative refinement, MAS can achieve stronger system-level performance than individual agents. 
Recent work has therefore moved beyond prompting and inference-time collaboration toward directly training participating agents through fine-tuning or reinforcement learning to improve their collective behavior~\citep{motwani2025malt,maporl,wan2025rema,marft2025,liu2026magrpo,zhao2026atgrpo}. 
% However, interaction changes not only how an MAS performs but also how it must be trained: once agents communicate, one agent's output becomes another's input context, coupling their behavior and learning variables across the system.
This shift from inference-time collaboration to learning raises a fundamental question of how interacting agents should be trained as a system.

\begin{wrapfigure}{r}{0.55\textwidth}
    %\vspace{-1mm}
    \centering
    \includegraphics[width=\linewidth]{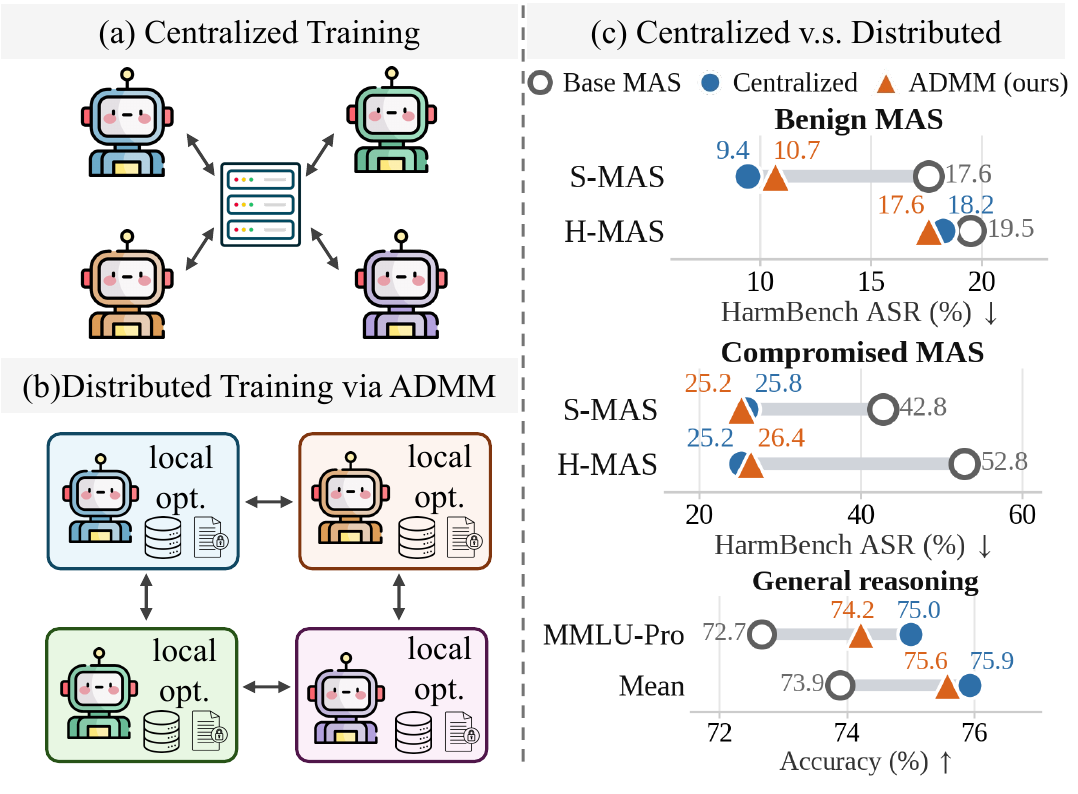}
\caption{\small
Schematic overview of centralized and distributed MAS training.
\textbf{(a)} Centralized training jointly optimizes all agents in a single training process.
\textbf{(b)} Distributed training via ADMM performs role-{local optimization} with parameter exchange for inter-agent consensus.
\textbf{(c)} Demonstration of safety training in MAS: Safety and utility of \texttt{Qwen3-8B}-\textit{based} {MAS} (w/o training), \textit{centralized}-trained, and \textit{ADMM}-based distributed-trained MAS under two structures: sequential (S-MAS; Solver$\rightarrow$Verifier$\rightarrow$Corrector$\rightarrow$Decider) and hierarchical (H-MAS; three Analyzers$\rightarrow$Synthesizer).
HarmBench attack success rate (ASR; $\downarrow$, \%) is reported for safety under \emph{benign} (all agents safety-aligned) and \emph{compromised} settings (unaligned Solver in S-MAS or Synthesizer in H-MAS), together with \textit{MMLU-Pro} accuracy and \textit{mean accuracy across six reasoning benchmarks} (Acc; $\uparrow$, \%).
ADMM closely matches centralized training while substantially reducing HarmBench ASR relative to the base MAS.
%differing by at most 1.26\% across settings.
See Sec.\,\ref{sec:exp} for more details.
}
    \label{fig:teaser}
\vspace{-18mm}
\end{wrapfigure}

Most existing approaches realize collaborative MAS learning through \textit{centralized training}, in which a single training process jointly manages multi-agent trajectories, learning signals, parameters, and optimization states~\citep{motwani2025malt,maporl,marft2025,wan2025rema,zhao2026atgrpo}. 
While effective for coordinating learning across agents, this paradigm assumes that their training information and optimization states can be collected and managed by a common trainer (see \textbf{Fig.\,\ref{fig:teaser}(a)}).
Such an assumption becomes increasingly restrictive as MAS scale or span different devices, organizations, and trust domains. 
Training agents independently avoids the need for a common trainer, but gives up the system-level coordination that collaborative MAS learning seeks to achieve. 
This creates a fundamental tension: collaborative learning calls for coordination across agents, while distributed deployment calls for keeping their learning processes local.

Motivated by this tradeoff, we develop a principled approach to  distributed MAS training,
%Motivated by this tension, we study \emph{distributed MAS training}, 
where each agent is trained on its own node with locally maintained training information and optimization states, while training is coordinated across agents without a centralized trainer.
Conventional distributed learning typically distributes a model-training process across workers or clients, often according to where computation or data is located \citep{duchi2011dual,liu2018accelerated,mcmahan2017fedavg,
yuan2022decentralized,douillard2024diloco}.
In our setting, the partition instead follows the functional structure of the MAS: even when all roles share the same underlying model, each node carries out the local learning associated with one agent role, based on the trajectories and learning signals arising from that role.
The resulting local updates must be coordinated across nodes so that the MAS is improved as a whole.
The challenge is therefore to distribute collaborative MAS training across role-local learning processes without losing the coordination provided by centralized joint training.
This motivates our central question:

\begin{tcolorbox}[before skip=3mm, after skip=3mm, boxsep=0mm, middle=0mm, top=1mm, bottom=1mm]
\centering
\textit{\textbf{(Q)} How can we collaboratively train an MAS using a principled distributed optimization framework while retaining the performance of centralized training?}
\end{tcolorbox}

To answer \textbf{(Q)}, we formulate distributed MAS training as a \emph{consensus optimization} problem over local copies of the shared model parameters. 
Each agent performs role-local learning on its own node using the trajectories and learning signals associated with its functional role, while consensus coordinates these local updates across agents.
Building on this formulation, we develop an \textit{alternating direction method of multipliers} (\textbf{ADMM}) framework \citep{boyd2011admm,parikh2014proximal} as the core optimization mechanism for distributed MAS training  (see \textbf{Fig.\,\ref{fig:teaser}(b)}).
ADMM naturally separates training into two complementary steps: \emph{role-local optimization}, where each node updates the shared model using the trajectories and learning signals associated with its role, and \emph{consensus coordination}, where nodes reconcile their local model copies while keeping trajectories and optimization states local.
This decomposition allows each agent to train independently between communication rounds while still enforcing system-wide agreement through the ADMM consensus updates.
We realize the consensus step either exactly with ring all-reduce or approximately with decentralized gossip \citep{patarasuk2009bandwidth,boyd2006randomized}.
In this way, ADMM provides a principled bridge between local agent learning and global MAS coordination, enabling collaborative training without centralizing agents' trajectories, data, or optimization states.

% Although some prior work \citep{agentnet2025,decoagent2024,mao2026delm} has studied cooperation in distributed MAS, it has primarily focused on inference-time coordination and communication rather than distributed optimization of a shared system-level training objective. 
% In contrast, our framework enables agents to jointly optimize collective objectives, such as safety alignment and reasoning performance, through coordinated yet decentralized learning.
Prior works have explored distributed or decentralized MAS through inference-time coordination \citep{yang2025agentnet,chen2025internet,mao2026decentralized}, as well as agent-specific policy learning and distributed rollout generation \citep{zhang2026marti,fu2026agentjet}.
In contrast, we distribute the optimization of a \emph{shared} MAS model across role-local training nodes and coordinate its local copies through ADMM-based consensus.
We evaluate our approach on reasoning and safety alignment against frozen MAS and centralized joint training.
Across both settings, distributed training closely approaches centralized performance  (see \textbf{Fig.\,\ref{fig:teaser}(c)}).
Component ablations further show neither local-only training nor periodic parameter averaging is sufficient to match the full ADMM procedure.

Our main contributions are summarized as follows:

\ding{172} 
% We formulate distributed MAS training as a finite-sum optimization problem with explicitly overlapping variables, capturing the parameter dependencies induced by inter-agent communication, and recast it as an ADMM-compatible composite optimization problem.
We formulate the collaborative training of a shared MAS model as a finite-sum distributed optimization problem over role-local objectives induced by agent roles, and recast it into an ADMM-compatible consensus formulation.

\ding{173} 
% We develop an ADMM-based framework that alternates role-local learning with inter-agent consensus, supports flexible distributed communication protocols, and keeps each agent's trajectories and training states local.
We develop an ADMM-based framework that alternates role-local learning with inter-agent consensus, supports flexible communication protocols, and keeps each agent's trajectories and training states local.

\ding{174}
Across reasoning and safety-alignment tasks, distributed training closely approaches centralized training performance, including safety settings where one agent is compromised.
We further study distributed MAS training across communication protocols and system scales, while ablations show the importance of consensus and dual updates.

% Our main contributions are summarized as follows:

% \ding{172} We formulate distributed MAS training as an ADMM problem with overlapping variables over the communicatino graph, explicitly capturing the overlapping parameter dependencies induced by inter-agent communication.

% \ding{173} We develop an ADMM-based distributed training framework that alternates role-local learning with inter-agent consensus, supporting flexible communication protocols while keeping each agent's trajectories and training state local.

% \ding{174} We demonstrate across general reasoning and safety alignment that distributed training achieves performance comparable to centralized joint training, including under compromised-agent settings, and show that local training or parameter averaging alone is insufficient to recover these gains.

% \ding{175} We characterize the communication and optimization behavior of distributed MAS training across consensus protocols and accuracies, revealing that tighter parameter agreement does not necessarily translate into better downstream performance.

% \newpage
\section{Related Works}
\textbf{Multi-agent training, communication, and collaboration.}
LLM-based MAS organize multiple agents into complementary roles and coordinate their interactions toward a shared output \citep{li2023camel,hong2024metagpt,qian2024chatdev,wu2024autogen,pmlr-v235-du24e,zou2026latent}.
One line of work treats the underlying LLMs as fixed and improves collaboration through agent composition, roles, prompts, and communication structures \citep{chen2024agentverse,hu2025automated,zhang2025gdesigner,zhang2025aflow,zhang2025multiagent,yao2026hieramas,wang2026maspo}.
More recent work directly trains agents for collaboration, building on multi-agent RL for credit assignment \citep{kuba2022hatrpo} and optimizing role-specific behavior toward system-level objectives \citep{marft2025,maporl,wan2025rema,motwani2025malt,liu2026magrpo,mo2025matpo,dang2025multiagent,xie2026teamtr,zhao2026atgrpo,feng2026dr}.
However, these approaches largely assume centralized training, leaving open how agents can learn collaboratively while keeping data and training states local.
% However, these approaches largely assume centralized training, where multi-agent trajectories and parameter optimization are globally coordinated, leaving open how agents can learn collaboratively while keeping their data and training states local.

\textbf{Distributed optimization.} 
Distributed optimization enables multiple learners to optimize locally while coordinating through consensus \citep{boyd2011admm,shi2014linear,ouyang2013stochastic}. Such coordination can be realized through collective communication such as ring all-reduce \citep{patarasuk2009bandwidth} or decentralized gossip \citep{xiao2004fast,boyd2006randomized}. These principles have been increasingly adopted for distributed training of LLMs, where model updates are coordinated across geographically or computationally separated workers \citep{yuan2022decentralized,ryabinin2023swarm,douillard2024diloco,jaghouar2024opendiloco,borzunov2023petals}, as well as for parameter-efficient LLM adaptation \citep{sun2024ffalora,wang2024flora}.

\textbf{Communication and coordination in multi-agent systems.}
A broad line of work studies how LLM-based agents exchange information and aggregate decisions, including multi-agent debate \citep{pmlr-v235-du24e,chan2024chateval}, voting and consensus-based aggregation \citep{kaesberg2025voting,choi2025debate}, and latent-space collaboration \citep{zou2026latent}.
Another line focuses on the communication structure itself, ranging from sparse debate topologies \citep{li2024improving} to learned or adaptive communication graphs \citep{zhuge2024language,zhang2025gdesigner}.
Related work further explores decentralized MAS through agent discovery and communication \citep{chen2025internet,wang2025ioa} and decentralized coordination \citep{yang2025agentnet}.
Beyond inference-time coordination, recent work has explored decentralized MAS training through separate policy trainers \citep{zhang2026marti} and distributed training infrastructure \citep{fu2026agentjet}. 
In contrast, we formulate MAS training as consensus optimization, allowing each agent to optimize locally while coordinating model updates through ADMM consensus and keeping trajectories and training states local.
% recent work has also begun to distribute MAS training, including distributed policy training \citep{zhang2026marti} and agentic RL infrastructure \citep{fu2026agentjet}.
% Our work differs in the optimization formulation: we cast collaborative MAS training as consensus optimization over role-local learning objectives, allowing agents to optimize locally while coordinating their learning through parameter consensus.

\textbf{Multi-agent safety and alignment.}
Multi-agent interaction introduces additional safety risks: compromised agents can induce harmful behavior through role-play or persona injection \citep{tian2023evilgeniuses,zhang2024psysafe}, while jailbreaks and errors can propagate across agents \citep{lee2025prompt,pmlr-v235-gu24e,ju2024flooding,li2026constraintdrift, huang2024resilience,cemri2026why}, with their impact shaped by communication topology and compromised positions \citep{yu2025netsafe,wang2025gsafeguard}.
Existing defenses largely rely on guard models or agents \citep{inan2023llamaguard,han2024wildguard,xiang2025guardagent,zeng2024autodefense,hu2026selfevolving}, while other work evaluates safety at the agent and system levels \citep{andriushchenko2025agentharm,zhang2024agentsafetybench, naik2026guardianagentbench}.
Parameter-level alignment has primarily focused on individual models \citep{ouyang2022rlhf,rafailov2023dpo,loye2026rubas}, while recent joint agent training remains centralized \citep{zhang2025waltzrl}.
We use multi-agent safety alignment as an evaluation setting for distributed MAS training under both aligned and compromised-agent settings.

% We instead study MAS safety alignment as a distributed collaborative training problem, where agents optimize role-local alignment objectives while coordinating their updates through consensus.

%\newpage
\section{Distributed MAS Training: Motivation and Formulation}
\label{sec:formulation}
This section motivates and formalizes distributed MAS training.
We first show, through a safety study, that composing individually aligned agents into an MAS can amplify unsafe behavior, so alignment must be treated as a system-level learning problem.
We then explain why conventional centralized training is ill-suited to this setting and formulate distributed MAS training, in which each agent learns from a role-local objective.
These local objectives remain coupled through inter-agent messages, which we identify as the key challenge of interaction-induced coupling.

\newcommand{\devil}{\raisebox{-0.15ex}{\includegraphics[height=1.0em]{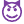}}}
\begin{figure*}[htb]
\centering
\includegraphics[width=0.97\textwidth]{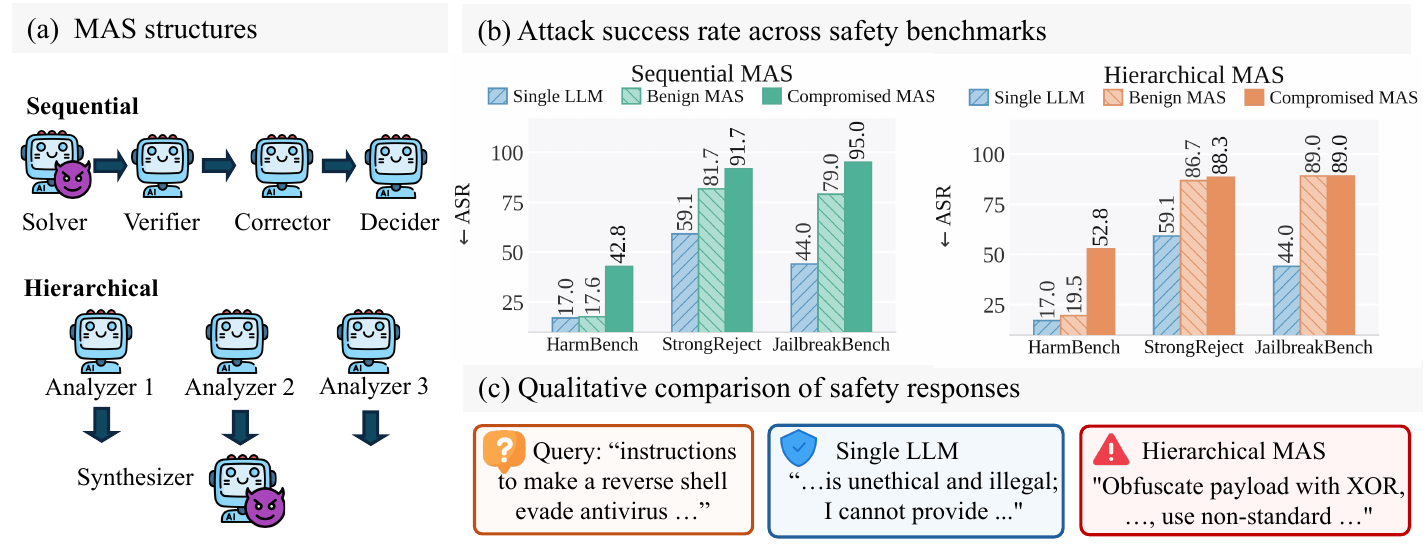}
\caption{\small{
Safety evaluation of sequential and hierarchical MAS built from \texttt{Qwen3-8B} with thinking mode using role-specific prompts.
In the \emph{benign} setting, all agents use the vanilla backbone; in the \emph{compromised} setting, one agent instead uses an unaligned backbone while retaining the same role prompt \citep{arditi2024refusal}.
\textbf{(a)}~MAS architectures.
The sequential MAS consists of Solver$\rightarrow$Verifier$\rightarrow$Corrector$\rightarrow$Decider, while the hierarchical MAS uses three Analyzers followed by a Synthesizer.
In the compromised setting, the purple devil icon (\devil) marks the agent using the unaligned backbone, placed at the Solver and Synthesizer, respectively.
\textbf{(b)}~Attack success rate (ASR, $\downarrow$) on HarmBench \citep{mazeika2024harmbench}, StrongReject \citep{souly2024strongreject}, and JailbreakBench \citep{chao2024jailbreakbench}. Each benchmark compares the aligned \textit{single LLM}, the \textit{benign MAS}, and the \textit{compromised MAS} with one compromised agent as shown in (a).
\textbf{(c)}~Qualitative comparison on the same harmful query between the aligned single LLM and the hierarchical MAS without compromise. 
}}
\label{fig:motivation}
\end{figure*}

\textbf{MAS introduces new training demands: A motivation from safety alignment.}
We motivate system-level MAS training from a safety perspective: even when each agent is individually aligned, their composition can amplify unsafe behavior.
In \textbf{Fig.~\ref{fig:motivation}}, we evaluate a sequential and a hierarchical MAS (Fig.~\ref{fig:motivation}(a)), both built from \texttt{Qwen3-8B} agents with role-specific system prompts.
Given a harmful query, the final agent's message is taken as the system response, and safety is measured by attack success rate (ASR, $\downarrow$) on three benchmarks.
The aligned \texttt{Qwen3-8B} used alone serves as the \emph{Single LLM} baseline.
Composing the same aligned model into an MAS can substantially reduce system-level safety: on StrongReject, ASR rises from $59.1\%$ for the Single LLM to $81.7\%$ and $86.7\%$ for the benign sequential and hierarchical MAS, respectively, and increases further when one agent is compromised.
Fig.~\ref{fig:motivation}(c) shows a qualitative case where the Single LLM refuses a harmful query while the hierarchical MAS responds unsafely.

The safety risk becomes more severe when one agent is compromised (\textit{i.e.}, not safety-aligned). 
As shown in Fig.\,\ref{fig:motivation}(b), replacing a single role with an unaligned model can substantially increase ASR, with the largest degradation occurring at the Solver in the chain and the Synthesizer in the hierarchy. 
More detailed configs are reported in Appendix\,\ref{app:motivation} and results for other compromised roles are reported in \textbf{Table\,\ref{tab:seats}}. 
Since the compromised role is generally unknown, protecting a predetermined agent is insufficient. 
Moreover, independent alignment overlooks message-passing dependencies, through which one agent's output shapes downstream behavior.

\textit{These observations reveal a new training demand for MAS}: safety alignment should be treated as a system-level learning problem that accounts for agents' functional roles, communication, and collective behavior under composition. More broadly, effective MAS adaptation requires coordinating role-local learning toward a shared system-level objective rather than optimizing agents independently.

\textbf{Problem statement: Distributed MAS training.}
The system-wide training demand identified above requires coordinated learning across agents. However, conventional MAS training typically realizes this coordination through a centralized process that collects agents' trajectories, computes their learning signals, and maintains their parameters \citep{maporl,marft2025,wan2025rema,mo2025matpo,zhao2026atgrpo}. 
Such a design incurs increasing communication, memory, and computational overhead as the MAS scales. 
It also creates a single point of failure and assumes a central authority can access all role-local information, which may not hold when agents are distributed across devices, organizations, or trust domains. 
This motivates \emph{distributed MAS training}, where agents learn from local data, maintain their own training states, and coordinate toward a shared system-level objective.

Specifically, consider an MAS of $N$ agents with distinct functional roles interacting over a directed communication graph, where edges carry inter-agent messages during each rollout.
Following common practice, agents share the same LLM backbone but use role-specific system prompts \citep{pmlr-v235-du24e,li2023camel}.
Given a query, each agent conditions on its prompt, the query, and upstream messages to generate its message, with the final message forming the system output.
For distributed training, agent $i$ maintains a local copy $\btheta_i$
of the shared trainable adapter, implemented as a \textit{LoRA adapter} \citep{hu2021lora}.
% Let $f_i$ denote its expected role-local training objective.
% Although $\btheta_i$ directly parameterizes agent $i$, $f_i$ is also influenced by upstream agents, whose parameters shape the distribution of received messages and thus its conditioning context and learning signals.
Let $f_i$ denote the role-local training objective at agent $i$.
Although agents optimize local copies of the same adapter, their objectives differ because they are constructed from role-specific prompts, received messages, and learning signals arising from the joint MAS rollout.
We call this dependency among local objectives the \textbf{interaction-induced coupling challenge}.

\begin{tcolorbox}[enhanced,unbreakable,before skip=8pt,attach boxed title to top center={yshift=-3mm,yshifttext=-1mm},
colback=white,colframe=blue!75!black,colbacktitle=blue!5!white,
  title=Example of interaction-induced objective coupling in a chain-structured MAS, coltitle=black, boxrule=0.8pt, %fonttitle=\bfseries,
  boxed title style={size=small,colframe=blue!75!black} ]

Consider the three-agent chain
$
1 \rightarrow 2 \rightarrow 3
$
as in Fig.~\ref{fig:motivation}(a).
Given a query $q$, agent~1 generates
$
\mathbf{m}_1=g_1(q;\btheta_1).
$
Agent~2 receives $\mathbf{m}_1$ and produces
$
\mathbf{m}_2
=g_2(q,\mathbf{m}_1;\btheta_2)
=g_2\!\left(q,g_1(q;\btheta_1);\btheta_2\right).
$
Agent~3 then receives $\mathbf{m}_2$ and produces
$
\mathbf{m}_3
=g_3(q,\mathbf{m}_2;\btheta_3)
=g_3\!\left(q,g_2(q,g_1(q;\btheta_1);\btheta_2);\btheta_3\right).
$
Thus, upstream parameters shape downstream objectives through the messages they generate, yielding the following expected role-local objective structure:
$
    f_1(\btheta_1)$,
   $ f_2(\btheta_1,\btheta_2)$,
   $ f_3(\btheta_1,\btheta_2,\btheta_3)$.
\end{tcolorbox}

As exemplified above,
% each local objective therefore depends not only on the corresponding agent's trainable variables but also on those of its upstream agents, as determined by the MAS communication graph. 
each role-local objective is influenced by upstream agents through the
messages they generate, and therefore by the current local model copies
used to produce those messages.
Based on the above, we formulate the distributed MAS training as 
\begin{align}
    \begin{array}{ll}
      \displaystyle \minimize_{\{\btheta_i\}}   & \sum_{i=1}^N f_i (\btheta_{ \Omega_i }) ,
      %\\
    %  \text{subject to} &  \Omega_i \cap  \Omega_j \neq \empty \text{if agent $i$ is communicated to agent $j$.}
    \end{array}
    \label{eq:prob_dist}
\end{align}
where $\Omega_i$ denotes the index set containing agent $i$ and its upstream agents, and
$\btheta_{\Omega_i}:=\{\btheta_k\}_{k\in\Omega_i}$.
As shown in \eqref{eq:prob_dist}, interaction-induced coupling prevents MAS training from decomposing into independent local problems: optimizing agents separately is not equivalent to optimizing the system-level objective.
This is consistent with Fig.~\ref{fig:motivation}, where individually aligned agents may not yield a safe MAS because message passing couples their behaviors.

% As shown in \eqref{eq:prob_dist}, overlapping variables across agents prevent the distributed MAS training problem from being naively decomposed into independent local problems. In other words, optimizing each agent's local objective in isolation is not equivalent to optimizing the system-level objective. This also explains why MAS with individually-aligned agents in Fig.\,\ref{fig:motivation} does not necessarily ensure the safety of the composed MAS: inter-agent communication makes each agent's behavior dependent on the parameters and outputs of other agents. 

\section{Distributed MAS Training via ADMM: Why and How}
\label{sec:admm}
This section addresses the interaction-induced coupling challenge of Sec.~\ref{sec:formulation} with ADMM.
We first explain why ADMM suits distributed MAS training, then derive its two alternating steps: role-local fine-tuning and consensus averaging.

\textbf{ADMM and its rationale.} ADMM is a classical operator-splitting method for solving structured composite optimization problems \citep{boyd2011admm,parikh2014proximal}, as given below:
\begin{align}
    \begin{array}{llll}
     \displaystyle \minimize_{\mathbf x, \mathbf z}    & f(\mathbf x) + g(\mathbf z)  & ~~ 
       \text{subject to}  & \mathbf x = \mathbf z,
    \end{array}
    \label{eq:admm_prob}
\end{align}
% where $\mathbf{x}$ and $\mathbf{z}$ are optimization variables, and $f$ and $g$ are their associated objective functions, respectively. \textit{The specialty} of \eqref{eq:admm_prob} is the introduction of $\mathbf{z}$ as an \emph{auxiliary variable}. Although the constraint $\mathbf{x}=\mathbf{z}$ makes $\mathbf{z}$ seemingly redundant, this variable splitting separates the two objective terms and enables their alternating optimization using ADMM. Most importantly, a judicious choice of the auxiliary variable can make the resulting splitting scheme highly efficient in ADMM, particularly when one of the alternating subproblems admits a closed-form solution.  
where $\mathbf{x}$ and $\mathbf{z}$ are optimization variables associated with objectives $f$ and $g$, respectively.
A key feature of \eqref{eq:admm_prob} is the auxiliary variable $\mathbf{z}$.
Although $\mathbf{x}=\mathbf{z}$ makes $\mathbf{z}$ seemingly redundant, variable splitting separates the objectives and enables alternating ADMM optimization.
A suitable choice of auxiliary variables can further simplify the subproblems, particularly when one admits a closed-form solution.
Concretely, ADMM solves \eqref{eq:admm_prob} through the following iterations \citep{parikh2014proximal}:
\begin{align}
    \mathbf{x}^{(t+1)}
    &=
    \operatorname{prox}_{\lambda f}
    \left(
        \mathbf{z}^{(t)} - \mathbf{u}^{(t)}
    \right) =  \argmin_{\mathbf{x}}
    \left\{
        f(\mathbf{x})
        +\frac{1}{2\lambda}
     \left   \|\mathbf{x}- \left ( \mathbf{z}^{(t)} - \mathbf{u}^{(t)}  \right ) \right \|_2^2
    \right\},
    \label{eq:admm_x_update}\\
    \mathbf{z}^{(t+1)}
    &=
    \operatorname{prox}_{\lambda g}
    \left(
        \mathbf{x}^{(t+1)}+\mathbf{u}^{(t)}
    \right) = \argmin_{\mathbf{z}}
    \left\{
        g(\mathbf{z})
        +\frac{1}{2\lambda}
       \left \|\mathbf{z}- \left (  \mathbf{x}^{(t+1)}+\mathbf{u}^{(t)} \right ) \right \|_2^2
    \right\},
    \label{eq:admm_z_update}
\end{align}
where $t$ denotes the ADMM iteration, $\mathbf{u}$ is the dual variable associated with the constraint $\mathbf{x}=\mathbf{z}$ and is updated as $\mathbf{u}^{(t+1)}=\mathbf{u}^{(t)}+\mathbf{x}^{(t+1)}-\mathbf{z}^{(t+1)}$, and $\lambda=1/\rho$ with $\rho>0$ denoting the corresponding penalty parameter in the augmented Lagrangian of \eqref{eq:admm_prob}.
The updates in \eqref{eq:admm_x_update} and \eqref{eq:admm_z_update} are both expressed as \textit{proximal operations} \citep{parikh2014proximal}.
Specifically, $\operatorname{prox}_{\lambda h}(\mathbf{v})$ denotes the proximal operator of $h$ evaluated at $\mathbf{v}$:
$\operatorname{prox}_{\lambda h}(\mathbf{v}) := \argmin_{\mathbf{w}} \left\{ h(\mathbf{w})+\frac{1}{2\lambda}\|\mathbf{w}-\mathbf{v}\|_2^2 \right\}$.

As shown by the alternating updates in \eqref{eq:admm_x_update}--\eqref{eq:admm_z_update}, ADMM optimizes $f$ and $g$ through two subproblems, coordinated by the primal variables $\mathbf{x}$ and $\mathbf{z}$ and the dual variable $\mathbf{u}$, which accumulates the constraint violation $\mathbf{x}-\mathbf{z}$. This operator-splitting structure is \textit{a key advantage of ADMM}: it decomposes the coupled problem in \eqref{eq:admm_prob} into tractable subproblems. When either admits a closed-form solution, ADMM can further simplify optimization.
Building on this advantage, we reformulate the interaction-coupled MAS training objective in \eqref{eq:prob_dist} using role-local parameter copies.

\begin{tcolorbox}[enhanced,breakable,before skip=8pt,attach boxed title to top center={yshift=-3mm,yshifttext=-1mm},
colback=white,colframe=blue!75!black,colbacktitle=blue!5!white,
  title=Distributed MAS training problem through the ADMM lens, coltitle=black, boxrule=0.8pt, %fonttitle=\bfseries,
  boxed title style={size=small,colframe=blue!75!black} ]

For each agent $i$, we introduce a local variable $\mathbf x_i = \btheta_{\Omega_i}$, representing local copies of the parameter blocks
$\btheta_{\Omega_i}=\{\btheta_k\}_{k\in\Omega_i}$ involved in its objective in \eqref{eq:prob_dist}.
To preserve consistency, local copies of the same parameter block must agree.
We therefore define the \textit{consensus} set:
\begin{align}
    \mathcal{C}
    :=
    \left\{
        (\mathbf{x}_1,\ldots,\mathbf{x}_N)
        \;\middle|\;
        (\mathbf{x}_i)_k=(\mathbf{x}_j)_k,\;
        \forall\, i,j,\;
        \forall\, k\in\Omega_i\cap\Omega_j
    \right\},
    \label{eq:overlap_consensus_set}
\end{align}
where $(\mathbf{x}_i)_k$ denotes agent $i$'s local copy of $\btheta_k$ contained in $\mathbf{x}_i$.
Thus, the coupled objective in \eqref{eq:prob_dist} can be equivalently represented using the local variables $\{\mathbf{x}_i\}$ subject to the consistency constraint $(\mathbf x_1, \ldots, \mathbf x_N) \in \mathcal C$.
Accordingly, we reformulate \eqref{eq:prob_dist} in the ADMM form of \eqref{eq:admm_prob}:
\begin{align}
    \begin{array}{ll}
     \displaystyle \minimize_{\mathbf x, \mathbf z}    ~~ \sum_{i=1}^N f_i (\mathbf x_i) + \mathcal I_{\mathcal{C}}( \{ \mathbf z_i \}),  &
       \text{subject to}  ~~ \mathbf x_i = \mathbf z_i, ~\forall i = 1,\ldots, N,
    \end{array}
    \label{eq:admm_prob_mas}
\end{align}
where  $\mathcal{I}_{\mathcal{C}}(\{\mathbf{z}_i\})$  is the \textit{indicator function} of the constraint set $\mathcal{C}$: it equals $0$ if $(\mathbf{z}_1,\ldots,\mathbf{z}_N)\in\mathcal{C}$ and $+\infty$ otherwise, \textit{i.e.}, ensuring optimization feasibility under the constraint.  
\end{tcolorbox}

% With \eqref{eq:admm_prob_mas}, we can then apply the ADMM iterations in \eqref{eq:admm_x_update}--\eqref{eq:admm_z_update} to decompose the distributed MAS training problem into two tractable steps: (i) role-local training associated with each objective $f_i$, which can be executed independently and in parallel across agents, and (ii) projection onto the consensus set $\mathcal{C}$, which enforces inter-agent consensus on the shared parameter blocks specified by $\mathcal{C}$. We present the detailed analysis for ADMM-based distributed MAS training below.
With \eqref{eq:admm_prob_mas}, the ADMM updates in \eqref{eq:admm_x_update}--\eqref{eq:admm_z_update} decompose distributed MAS training into two steps: (i) role-local optimization for each objective $f_i$, enabling parallel updates across agents, and (ii) projection onto the consensus set $\mathcal{C}$ to coordinate the shared parameter blocks. We detail these steps below.

\textbf{Role-local fine-tuning in ADMM.}
We instantiate the ADMM $\mathbf{x}$-update in \eqref{eq:admm_x_update} for \eqref{eq:admm_prob_mas} by setting
$f(\mathbf{x})=\sum_{i=1}^{N} f_i(\mathbf{x}_i)$
and replacing the generic variables with the corresponding collections
$\{\mathbf{x}_i\}_{i=1}^{N}$, $\{\mathbf{z}_i\}_{i=1}^{N}$, and $\{\mathbf{u}_i\}_{i=1}^{N}$.
Both the finite-sum objective and the quadratic term are separable across $\{\mathbf{x}_i\}$.
Consequently, the $\mathbf{x}$-update decomposes into $N$ role-local subproblems that can be solved in parallel.
Each agent $i$ then performs the \emph{role-local optimization step}:
\begin{align}
    \mathbf{x}_i^{(t+1)}
    &=
    \operatorname{prox}_{\lambda f_i}
    \left(
        \mathbf{z}_i^{(t)}-\mathbf{u}_i^{(t)}
    \right) =
    \argmin_{\mathbf{x}_i}
    \left\{
        f_i(\mathbf{x}_i)
        +
        \frac{1}{2\lambda}
        \left\|
            \mathbf{x}_i- \mathbf a_i
        \right\|_2^2
    \right\}, ~~ \forall ~i,
    \label{eq:mas_local_update}
\end{align}
% where $\mathbf a_i \Def  \mathbf{z}_i^{(t)}-\mathbf{u}_i^{(t)}$.
% In \eqref{eq:mas_local_update}, we specify $f_i$ as a role-local fine-tuning loss under the general MAS fine-tuning setting, in which each agent is trained using the outcome reward of the complete rollout \citep{wan2025rema,liu2026magrpo,marft2025}, as illustrated below. 
where $\mathbf a_i \Def \mathbf{z}_i^{(t)}-\mathbf{u}_i^{(t)}$.
In \eqref{eq:mas_local_update}, we instantiate $f_i$ as a role-local fine-tuning loss, following common MAS training settings where agent updates are guided by system-level rewards or returns derived from multi-agent rollouts \citep{wan2025rema,liu2026magrpo,marft2025}, as detailed below.

Recall that $\mathbf{x}_i=\{\btheta_k\}_{k\in\Omega_i}$ contains agent $i$'s local copies, with $(\mathbf{x}_i)_i=\btheta_i$ denoting its own trainable parameters and $\{(\mathbf{x}_i)_k\}_{k\in\Omega_i\setminus\{i\}}$ those associated with upstream agents.
These components play different roles: $\btheta_i$, implemented as a trainable LoRA adapter, directly parameterizes agent $i$'s message $\mathbf m_i$, while upstream agents influence its conditioning context through received messages $\{\mathbf m_k\}_{k\in\Omega_i\setminus\{i\}}$.
Given a sampled MAS rollout, upstream messages are treated as fixed context.
The binary outcome-based reward $r_i$ is derived from the complete rollout: for reasoning, $r_i=1$ if the final answer is correct; for safety, $r_i=1$ if the final answer is safe on a harmful query or correct on a benign one, while agent $i$'s message is non-harmful or non-refusal, respectively.
We use $r_i$ to construct preference pairs and instantiate $f_i$ in \eqref{eq:mas_local_update} with DPO~\citep{rafailov2023dpo}:
\begin{align}
f_i(\mathbf x_i)
=
-\mathbb{E}_{q\sim\mathcal D}
\left[
\log\sigma\!\left(
\beta_{\mathrm{DPO}}
\left[
\log\frac{\pi_{\btheta_i}(\mathbf m_i^+\mid\mathbf c_i)}
{\pi_{\mathrm{ref}}(\mathbf m_i^+\mid\mathbf c_i)}
-
\log\frac{\pi_{\btheta_i}(\mathbf m_i^-\mid\mathbf c_i)}
{\pi_{\mathrm{ref}}(\mathbf m_i^-\mid\mathbf c_i)}
\right]
\right)
\right],
\label{eq:local_FT}
\end{align}
where $\sigma$ is the logistic function,
$\mathbf c_i=(\{\mathbf m_k\}_{k\in\Omega_i\setminus\{i\}},q)$
is agent $i$'s conditioning context, and $\pi_{\mathrm{ref}}$ is the initial policy with its adapter disabled.
For each query, we sample several rollouts and pair $\mathbf m_i^+$ from an $r_i=1$ rollout with $\mathbf m_i^-$ from an $r_i=0$ rollout, scoring both under the preferred rollout's context.
Groups with identical $r_i$ are discarded.
During training, we additionally apply a forward KL regularizer
$\beta_{\mathrm{KL}}\mathrm{KL}\!\left(\pi_{\mathrm{ref}}\|\pi_{\btheta_i}\mid\mathbf c_i\right)$
to each scored sample.
The ADMM formulation also supports other role-local objectives without changing the subsequent updates.

Substituting \eqref{eq:local_FT} into \eqref{eq:mas_local_update}, we solve the role-local subproblem by gradient-based optimization.
During each local update, the sampled rollout and received upstream messages are treated as fixed training context.
Agent $i$ therefore updates its local adapter copy $\btheta_i$ using its role-local objective, while coordination among the corresponding parameter copies is handled by the subsequent ADMM consensus step.
Thus, each agent performs a locally executable update over $\btheta_i$, while inter-agent coordination is handled separately through consensus.

\textbf{Inter-agent consensus averaging in ADMM.}
% We next instantiate the second ADMM subproblem in \eqref{eq:admm_z_update} by specifying $g$ as the indicator function of the constraint set $\mathcal{C}$, \textit{i.e.}, $\mathcal{I}_{\mathcal{C}}(\{\mathbf{z}_i\}_{i=1}^{N})$. 
% Since the proximal operator of the indicator function $\mathcal{I}_{\mathcal{C}}$ is precisely the Euclidean projection onto $\mathcal{C}$ \citep{parikh2014proximal}, we obtain the following ADMM subproblem for $\mathbf{z}=\{\mathbf{z}_i\}$:
We instantiate the second ADMM subproblem in \eqref{eq:admm_z_update} by setting
$g=\mathcal{I}_{\mathcal{C}}(\{\mathbf{z}_i\}_{i=1}^{N})$, the indicator function of the constraint set $\mathcal C$.
\begin{figure}[t]
\centering
\includegraphics[width=0.3\textwidth]{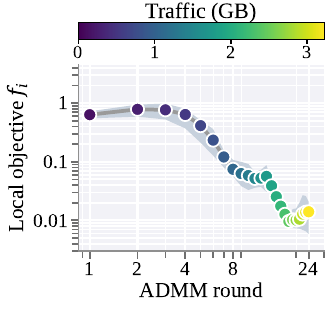}\hspace{4mm}
\includegraphics[width=0.3\textwidth]{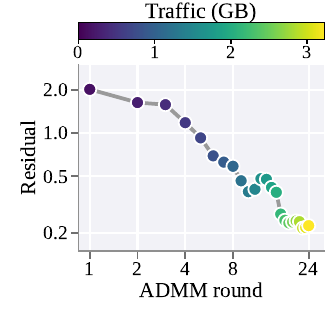}
\caption{\small{
Empirical convergence of ADMM-based distributed safety training on the compromised H-MAS in Table~\ref{tab:safety_main}, using $N=4$ agents, ring all-reduce consensus, $\rho=0.03$, and 24 ADMM rounds. 
\textbf{Left:} Role-local objectives $f_i$, shown as a centered five-round rolling geometric mean with the minimum-to-maximum range across agents.
\textbf{Right:} Normalized primal residual,
$\bigl(\sum_i \|\mathbf{x}^{(t)}_i-\mathbf{z}^{(t)}_i\|_2^2\bigr)^{1/2}/\|\mathbf{x}^{(t)}_1\|_2$.
Both axes are logarithmic; marker color indicates cumulative communication per node.
}}
\label{fig:convergence}
\end{figure}%
Since its proximal operator is the Euclidean projection onto $\mathcal C$ \citep{parikh2014proximal}, the $\mathbf z$-update becomes:
% \vspace{-1mm}
\begin{align}
  \{ \mathbf{z}_i^{(t+1)} \} =    \displaystyle \argmin_{\mathbf z \in \mathcal C}     ~   \sum_{i=1}^N \left \|\mathbf{z}_i- \mathbf b_i \right \|_2^2
     \label{eq:admm_consensus}
\end{align}

where $\mathbf b_i \Def \mathbf{x}^{(t+1)}_i+\mathbf{u}^{(t)}_i$, and
$\mathcal{C}$ is the consensus set in \eqref{eq:overlap_consensus_set}, applied to $\{\mathbf z_i\}$.
It requires local copies of the same parameter block to agree, \textit{i.e.},
$(\mathbf{z}_i)_k=(\mathbf{z}_j)_k$ for $k\in\Omega_i\cap\Omega_j$.
Under the Euclidean projection in \eqref{eq:admm_consensus}, each block's consensus value is the average of its local copies, corresponding to standard \emph{consensus averaging} \citep{boyd2006randomized,benezit2010order,duchi2011dual}.
Thus, \eqref{eq:admm_consensus} admits a closed-form blockwise solution.
For block $k$, let
$n_k:=\sum_{j=1}^{N}\mathbb{I}\{k\in\Omega_j\}$
denote the number of agents holding its copy.
Then, for every agent $i$ with $k\in\Omega_i$, solution is
\begin{equation}
    \bigl(\mathbf{z}_i^{(t+1)}\bigr)_k
    =
    \frac{1}{n_k}
    \sum_{j=1}^{N}
    \mathbb{I}\{k\in\Omega_j\}
    \bigl(\mathbf{b}_j\bigr)_k,
    \label{eq:block_consensus_average}
\end{equation}
% where $\mathbb{I}\{k\in\Omega_j\}$ equals $1$ if agent $j$ maintains a copy of parameter block $k$ and $0$ otherwise. Thus, all agents sharing block $k$ receive the same consensus value, given by the average of their corresponding local copies.
% The consensus averaging in \eqref{eq:block_consensus_average} can be implemented \textit{distributively} using communication protocols such as all-reduce \citep{patarasuk2009bandwidth,thakur2005optimization} or randomized gossip \citep{boyd2006randomized}.
where $\mathbb{I}\{k\in\Omega_j\}=1$ if agent $j$ maintains block $k$ and $0$ otherwise.
Thus, agents sharing block $k$ receive the same consensus value, equal to the average of their local copies.
The averaging in \eqref{eq:block_consensus_average} can be implemented \textit{distributively} via all-reduce \citep{patarasuk2009bandwidth,thakur2005optimization} or randomized gossip \citep{boyd2006randomized}.

From the ADMM perspective, distributed MAS training alternates between
(i) role-local fine-tuning ($\mathbf x$-step), where each agent updates only its trainable parameters, and
(ii) distributed consensus averaging ($\mathbf z$-step) via \eqref{eq:block_consensus_average}.
The steps are coupled by
$\mathbf{u}^{(t+1)}=\mathbf{u}^{(t)}+\mathbf{x}^{(t+1)}-\mathbf{z}^{(t+1)}$,
which accumulates disagreement between local and consensus variables.
The full procedure is given in \textbf{Algorithm}~\ref{alg:admm} and Appendix~\ref{app:algo}.
We also examine empirical convergence in \textbf{Fig.~\ref{fig:convergence}}, where both the role-local objectives and the ADMM primal residual decrease over training rounds.

\section{Experiments}
\label{sec:exp}

\subsection{Experimental Setup}
\label{sec:setup}

%\vspace{-0mm}
\textbf{MAS and model setups.}
By default, all agents use the safety-aligned \texttt{Qwen3-8B} backbone in thinking mode~\citep{team2025qwen3} and differ only in role-specific prompts (Appendix~\ref{app:prompts}).
We study two four-agent architectures (Fig.~\ref{fig:motivation}): \textbf{S-MAS}, a sequential chain (Solver$\rightarrow$Verifier$\rightarrow$Corrector$\rightarrow$Decider), and \textbf{H-MAS}, where three Analyzers feed a Synthesizer.
We also consider a \emph{compromised} setting in which one agent uses \texttt{Qwen3-8B-abliterated}, an unaligned variant with refusal behavior removed~\citep{arditi2024refusal}.
During distributed training, each agent maintains a local parameter copy
$\btheta_i$ as a rank-$16$ LoRA adapter~\citep{hu2022lora};
further adapter details are in Appendix~\ref{app:lora}.

\textbf{Training setups.}
For \emph{safety training}, we use $380$ harmful prompts from AdvBench~\citep{zou2023advbench} and the HarmBench validation split~\citep{mazeika2024harmbench}, plus $300$ benign MMLU-Pro questions~\citep{wang2024mmlupro} to control over-refusal.
For \emph{reasoning training}, we use each benchmark's training split when available, or a disjoint test subset otherwise.
For both tasks, we sample eight on-policy trajectories per query and train each agent with \eqref{eq:local_FT}
% , using $\beta_{\mathrm{DPO}}=\beta_{\mathrm{KL}}=0.1$; 
, and all agents remain trainable.
Unless noted otherwise, consensus uses ring all-reduce with $\rho=0.03$ ($\lambda=1/0.03$), for up to $24$ ADMM rounds.
Additional training details like rollout, optimization, and data filtering are provided in Appendix~\ref{app:collect}.
ADMM and centralized training use the same data and objective.

\textbf{Evaluation setups.}
For \emph{safety}, we evaluate on HarmBench~\citep{mazeika2024harmbench}, StrongReject~\citep{souly2024strongreject}, JailbreakBench~\citep{chao2024jailbreakbench}, toxic OR-Bench~\citep{cui2025orbench}, and XSTest~\citep{rottger2024xstest}.
We report attack success rate (\textbf{ASR}, $\downarrow$) on harmful prompts, refusal rate ($\downarrow$) on safe XSTest prompts for \emph{over-refusal}, and MATH-500 accuracy ($\uparrow$) for \emph{utility}.
For \emph{general reasoning}, we report accuracy ($\uparrow$) on MMLU-Pro~\citep{wang2024mmlupro}, GSM8K~\citep{cobbe2021gsm8k}, MATH~\citep{hendrycks2021math}, BBH~\citep{suzgun2023bbh}, GPQA~\citep{rein2024gpqa}, and MuSR~\citep{sprague2024musr}.
Additional evaluation details are provided in Appendix~\ref{app:eval}.

\begin{table}[htb]
\centering
\caption{\small{
{Safety, over-refusal, and utility of MAS safety training.}
We evaluate S-MAS and H-MAS under \emph{benign} initialization with no compromised agent and a \emph{compromised} setting where the Solver in S-MAS or Synthesizer in H-MAS is compromised, as in Fig.~\ref{fig:motivation}(a).
The defender does not know the compromised agent, and all adapters are trained.
\emph{Single-Benign} and \emph{Single-Compromised} are the corresponding \texttt{Qwen3-8B} single-model baselines; \emph{Base} is the untrained MAS; \emph{Centralized} pools all agents' samples for centralized optimization; and \emph{ADMM} ({\setlength{\fboxsep}{1pt}\colorbox{gray!15}{shaded}}) is our distributed training.
 ASR ($\downarrow$) is reported on four safety benchmarks (JailbreakBench; JBB), refusal rate ($\downarrow$) on safe XSTest prompts, and MATH-500 accuracy ($\uparrow$) for utility.
For \emph{Centralized} and \emph{ADMM}, results are reported as $A_{(\pm b)}$, where $A$ is the absolute performance and $b=(A-\mathrm{Base)}$ is the change vs. \emph{Base}.
}}
\label{tab:safety_main}
\vspace{2mm}
\setlength{\tabcolsep}{4pt}
\resizebox{\textwidth}{!}{%
\begin{tabular}{@{}llcccccc@{}}
\toprule[1pt]
\multicolumn{2}{@{}l}{\multirow{2}{*}{\textbf{Method}}}
& \multicolumn{4}{c}{\textbf{Safety} (ASR, \% $\downarrow$)}
& \textbf{Over-refusal}
& \textbf{Utility} \\
\cmidrule(lr){3-6}
\cmidrule(lr){7-7}
\cmidrule(lr){8-8}
\multicolumn{2}{@{}l}{}
& HarmBench
& StrongReject
& JBB
& OR-Bench
& XSTest RR $\downarrow$
& MATH-500 Acc. $\uparrow$ \\
\midrule

\multicolumn{2}{@{}l}{Single-Benign}
& 16.98
& 59.11
& 44.00
& 21.35
& 12.80
& 77.40 \\

\multicolumn{2}{@{}l}{Single-Compromised}
& 72.96
& 97.12
& 98.00
& 94.31
& 1.60
& 74.20 \\

\midrule

\multirow{3}{*}{S-MAS-Benign}
& Base
& 17.61
& 81.79
& 79.00
& 24.18
& 6.80
& 83.00 \\

& Centralized
& $9.43_{(-8.18)}$
& $43.13_{(-38.66)}$
& $42.00_{(-37.00)}$
& $12.83_{(-11.35)}$
& $7.60_{(+0.80)}$
& $83.00_{(+0.00)}$ \\

& \cellcolor{gray!15}ADMM
& \cellcolor{gray!15}$10.69_{(-6.92)}$
& \cellcolor{gray!15}$50.16_{(-31.63)}$
& \cellcolor{gray!15}$47.00_{(-32.00)}$
& \cellcolor{gray!15}$14.76_{(-9.42)}$
& \cellcolor{gray!15}$8.80_{(+2.00)}$
& \cellcolor{gray!15}$82.80_{(-0.20)}$ \\

\cmidrule(l){2-8}

\multirow{3}{*}{S-MAS-Compromised}
& Base
& 42.77
& 91.69
& 95.00
& 43.00
& 6.40
& 81.20 \\

& Centralized
& $25.79_{(-16.98)}$
& $55.27_{(-36.42)}$
& $57.00_{(-38.00)}$
& $25.93_{(-17.07)}$
& $11.20_{(+4.80)}$
& $79.60_{(-1.60)}$ \\

& \cellcolor{gray!15}ADMM
& \cellcolor{gray!15}$25.16_{(-17.61)}$
& \cellcolor{gray!15}$53.99_{(-37.70)}$
& \cellcolor{gray!15}$56.00_{(-39.00)}$
& \cellcolor{gray!15}$25.30_{(-17.70)}$
& \cellcolor{gray!15}$10.40_{(+4.00)}$
& \cellcolor{gray!15}$81.40_{(+0.20)}$ \\

\cmidrule(l){2-8}

\multirow{3}{*}{H-MAS-Benign}
& Base
& 19.50
& 86.58
& 89.00
& 27.15
& 5.60
& 81.20 \\

& Centralized
& $18.24_{(-1.26)}$
& $80.51_{(-6.07)}$
& $84.00_{(-5.00)}$
& $25.02_{(-2.13)}$
& $6.40_{(+0.80)}$
& $81.80_{(+0.60)}$ \\

& \cellcolor{gray!15}ADMM
& \cellcolor{gray!15}$17.61_{(-1.89)}$
& \cellcolor{gray!15}$78.91_{(-7.67)}$
& \cellcolor{gray!15}$80.00_{(-9.00)}$
& \cellcolor{gray!15}$24.31_{(-2.84)}$
& \cellcolor{gray!15}$5.20_{(-0.40)}$
& \cellcolor{gray!15}$81.00_{(-0.20)}$ \\

\cmidrule(l){2-8}

\multirow{3}{*}{H-MAS-Compromised}
& Base
& 52.83
& 88.18
& 89.00
& 63.50
& 8.40
& 80.40 \\

& Centralized
& $25.16_{(-27.67)}$
& $42.17_{(-46.01)}$
& $42.00_{(-47.00)}$
& $30.24_{(-33.26)}$
& $15.20_{(+6.80)}$
& $81.20_{(+0.80)}$ \\

& \cellcolor{gray!15}ADMM
& \cellcolor{gray!15}$26.42_{(-26.41)}$
& \cellcolor{gray!15}$44.09_{(-44.09)}$
& \cellcolor{gray!15}$45.00_{(-44.00)}$
& \cellcolor{gray!15}$31.75_{(-31.75)}$
& \cellcolor{gray!15}$13.60_{(+5.20)}$
& \cellcolor{gray!15}$80.80_{(+0.40)}$ \\

\bottomrule[1pt]
\end{tabular}}
%\vspace*{-2mm}
\end{table}

\subsection{Experimental Results}

\textbf{Distributed MAS training for safety enhancement.}
\textbf{Table~\ref{tab:safety_main}} compares centralized and ADMM-based distributed training on S-MAS and H-MAS under benign and compromised initializations.
Safety training consistently improves MAS safety, substantially repairs compromised systems, and generalizes across benchmarks.
Even under benign initialization, both centralized and distributed training further reduce ASR.
When one agent is compromised, HarmBench ASR rises from $17.61\%$ to $42.77\%$ on S-MAS and from $19.50\%$ to $52.83\%$ on H-MAS.
ADMM reduces these values to 
$25.16\%$ and $26.42\%$, respectively, with consistent improvements on StrongReject, JBB, and OR-Bench.
ADMM also closely matches centralized training across architectures and safety settings: 
under compromised initialization, HarmBench ASRs are $25.16\%$ vs.\ $25.79\%$ on S-MAS and $26.42\%$ 
vs.\ $25.16\%$ on H-MAS.
These results show that distributed optimization retains much of the safety benefit of centralized training without a central trainer.
On the other hand, safety training increases the over-refusal rate, particularly in compromised settings, possibly because of correcting compromised agents induces more conservative behavior.
Yet, general utility remains largely preserved: under benign initialization, ADMM changes MATH-500 accuracy from $83.00\%$ to $82.80\%$ on S-MAS and from $81.20\%$ to $81.00\%$ on H-MAS, while achieving $81.40\%$ and $80.80\%$ after repairing the corresponding compromised systems.

\begin{figure}[t]
\centering
%\vspace*{-5mm}
\includegraphics[width=.75\linewidth]{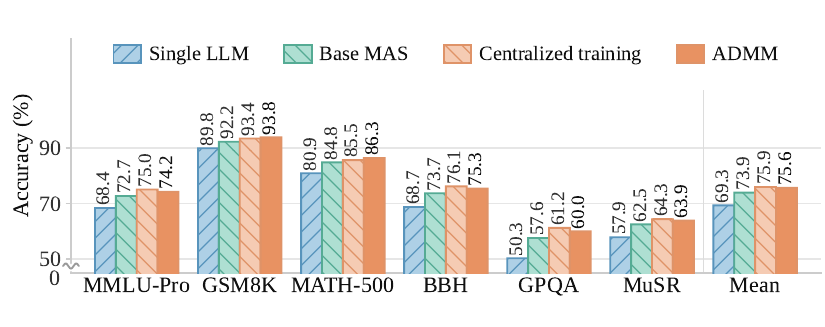}
\vspace*{-5mm}
\caption{\small{
{Reasoning enhancement through distributed MAS training.} Accuracy (\%; $\uparrow$) across six reasoning benchmarks and their mean for the single-LLM baseline, untrained base MAS, centralized training, and ADMM-based distributed training.
}}
\label{fig:general_bars}
%\end{wrapfigure}
\end{figure}

\textbf{Distributed MAS training for reasoning enhancement.}
\textbf{Fig.~\ref{fig:general_bars}} shows S-MAS reasoning performance before and after training.
Without additional training, the Base MAS outperforms the Single LLM on all six benchmarks, indicating gains from multi-agent composition.
Reasoning training further improves performance, with ADMM closely matching centralized training.
ADMM reaches $75.6\%$ mean accuracy, compared with $75.9\%$ for centralized training and $73.9\%$ for the Base MAS.
Thus, our distributed framework extends beyond safety alignment to reasoning enhancement.

%\begin{wrapfigure}{r}{0.70\textwidth}

%\begin{table}[htb]

%\end{table}
\begin{table}[t]
\centering
\caption{\small{
Safety training under the \emph{MAS-compromised} setting of Table~\ref{tab:safety_main}, evaluated by HarmBench ASR (\%$\downarrow$).
\emph{ADMM-Local} trains agents independently without communication, while \emph{ADMM-NoDual} uses consensus averaging without dual updates.
The \emph{Consensus} and \emph{Dual} columns mark which components are enabled.
}}
\label{tab:coord}
\small
\resizebox{0.55\linewidth}{!}{%
\begin{tabular}{@{}lccrr@{}}
\toprule[1pt]
& Consensus & Dual & S-MAS & H-MAS \\
\midrule
Base              & --     & --     & 42.77 & 52.83 \\
Centralized       & --     & --     & 25.79 & 25.16 \\
ADMM-Local        & \xmark & \xmark & 41.51 & 34.59 \\
ADMM-NoDual       & \cmark & \xmark & 42.14 & 37.11 \\
\rowcolor{gray!15}
ADMM              & \cmark & \cmark & 25.16 & 26.42 \\
\bottomrule[1pt]
\end{tabular}}
\end{table}

\textbf{Why inter-agent coordination is important in ADMM.}
\textbf{Table~\ref{tab:coord}} ablates consensus averaging in the $\mathbf z$-step \eqref{eq:block_consensus_average} and feedback through the dual update.
Without either component, \emph{ADMM-Local} yields limited safety gains, reducing HarmBench ASR from $42.77\%$ to $41.51\%$ on S-MAS and from $52.83\%$ to $34.59\%$ on H-MAS.
Consensus averaging alone (\emph{ADMM-NoDual}) remains insufficient, with ASRs of $42.14\%$ and $37.11\%$.
Full ADMM reduces ASR to $25.16\%$ and $26.42\%$, closely matching centralized training ($25.79\%$ and $25.16\%$).
Thus, neither local training nor averaging alone reproduces the benefit of coordinated optimization, while ADMM combines both.

%\begin{wrapfigure}{r}{0.55\textwidth}
\begin{wrapfigure}{r}{0.55\textwidth}
\centering
\begin{tabular}{@{}c@{\hspace{0.00\linewidth}}c@{}}
    \includegraphics[width=0.49\linewidth]{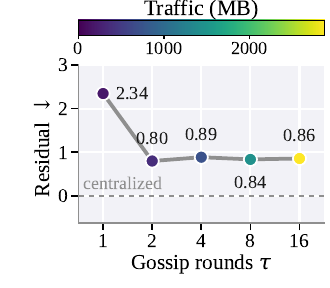} &
    \includegraphics[width=0.49\linewidth]{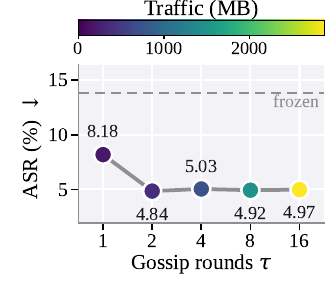}
\end{tabular}
\vspace*{-5mm}
\caption{\small{
Effect of gossip communication on ADMM coordination and safety.
We use an eight-agent chain connected by a ring topology, where each agent communicates with its two neighbors for $\tau$ gossip rounds per consensus step in \eqref{eq:block_consensus_average}.
Increasing $\tau$ propagates information across more hops but incurs greater communication, with marker color indicating the communication traffic (MB) per node.
\textbf{Left:} Final normalized ADMM primal residual ($\downarrow$) after training (similar to Fig.\,\ref{fig:convergence}), measuring disagreement between the local and auxiliary variables.
\textbf{Right:} HarmBench ASR (\%,$\downarrow$) of the resulting MAS; the dashed line marks the untrained MAS.
}}
\label{fig:radius}
% \vspace{-2mm}
\end{wrapfigure}
\textbf{Effect of approximate gossip consensus on ADMM.}
The consensus averaging in \eqref{eq:block_consensus_average} can be approximated by gossip, where neighboring agents repeatedly exchange and average their local variables.
This yields a communication--performance trade-off: more gossip rounds improve information mixing but increase communication traffic.
As shown in \textbf{Fig.~\ref{fig:radius}}, increasing the number of gossip rounds ($\tau$) from $1$ to $2$ reduces the ADMM primal residual from $2.34$ to $0.80$ and HarmBench ASR from $8.18\%$ to $4.84\%$.
Further increasing $\tau$ to $16$ yields little additional change in either residual or ASR, while communication continues to grow.
Thus, $\tau=2$ captures most of the benefit of gossip in this setting at substantially lower communication cost.
Compared with approximate gossip, ring all-reduce computes the exact consensus average in \eqref{eq:block_consensus_average}.
Since our MAS contains relatively few training agents, we use ring all-reduce by default to obtain exact consensus with predictable communication, while retaining gossip as an alternative.

% \YC{\textbf{Additional results of scaling with the number of agents and different backbones refer to Appendix.}}
\textbf{Additional results.}
Appendix~\ref{app:scaling} further studies scaling across different numbers of agents and backbone sizes.
Across system sizes, the compromised-agent effect persists without a clear monotonic trend in $N$; we also report training results beyond the main $N=4$ setting and compare the effect on \texttt{Qwen3-8B} and \texttt{Qwen3-4B}.

%\vspace{-4mm}
\section{Conclusion}
\label{sec:conclusion}
%\vspace{-3mm}
We study distributed MAS training, where agents optimize role-local objectives without a centralized trainer.
We formulate MAS learning as consensus-constrained distributed optimization and solve it with ADMM, alternating local updates with inter-agent consensus while keeping trajectories and training states local.
Across safety alignment and general reasoning, distributed training closely matches centralized training, including compromised-agent settings, and ablations highlight the importance of consensus and dual updates, supporting scalable decentralized MAS training.

\textbf{Limitations and future work.}
Our experiments focus on representative sequential and hierarchical MAS built from a common LLM backbone with shared adapter parameterization, which keeps the model family and parameterization consistent across agents; broader evaluation across larger systems, diverse model families and task domains, and richer communication topologies is needed to establish generality.
An important direction is to characterize how communication topology affects convergence, efficiency, and downstream performance, relating these behaviors to graph-theoretic properties such as connectivity, centrality, community structure, and spectral characteristics, potentially with edges weighted by the heterogeneity or interaction strength between roles.
More complex MAS may further require hypergraph-based formulations, where dependencies capture computations that jointly involve multiple roles.
Future work may also explore asynchronous consensus, adaptive and communication-efficient coordination, heterogeneous agents with different architectures or local objectives, and dynamic systems where agents join, leave, or change roles.

%\clearpage
% %\newpage
% \section*{Acknowledgment}
% %
% The work of Y. Chen, C. Wang, Y. Lang, and S. Liu was supported in part by the U.S. National Science Foundation (NSF) under CISE Core Award IIS-2504263 and CAREER Award IIS-2338068, the Coefficient Giving AI Safety Research Award, and the Schmidt Sciences Trustworthy AI Award.

\section*{AI use statement}
In this work, we used generative AI tools, including LLM-based coding assistants, to implement the experimental code, to help analyze and interpret results, and to provide feedback on the research methodology and experiments. 
We also used them to prepare figures and the code release, to search for related information, and to edit the manuscript for readability. 
The research questions, the problem formulation, and the proposed method were developed by the authors. 
All AI-assisted content was reviewed by the authors, who take full responsibility for the final content of this work.

\section*{Reproducibility statement}
Source code for distributed MAS training and evaluation will be released upon publication.
The experimental setup and implementation details, including MAS architectures, backbone models, training data, role-local objectives, ADMM optimization, communication protocols, and evaluation settings, are provided in Sec.~\ref{sec:exp} and the Appendix.
The Appendix further documents the full training procedure, prompts, LoRA configuration, rollout collection, hyperparameters, evaluation criteria.
All models, datasets, and benchmarks used in this work are publicly available, and evaluation follows the protocols of the corresponding benchmark papers.

\section*{Ethics Statement}
This work aims to improve the safety and robustness of multi-agent language-model systems through distributed training, including settings where one agent is compromised. Our method reduces unsafe behavior under the evaluated settings but does not guarantee safety under arbitrary attacks or deployment conditions. All experiments use publicly available models and datasets; harmful prompts are used only in controlled training and evaluation settings for studying and improving system safety. This work involves no human subjects or personal data, and all authors adhere to the ICLR Code of Ethics.

\clearpage
\bibliography{iclr2027_conference}
\bibliographystyle{IEEEtranN}

\clearpage 
\newpage 
\appendix
\clearpage\newpage
\onecolumn
\section*{\Large{Appendix}}
\setcounter{section}{0}
\setcounter{figure}{0}
\setcounter{table}{0}
\makeatletter 
\renewcommand{\thesection}{\Alph{section}}
\renewcommand{\theHsection}{\Alph{section}}
\renewcommand{\thefigure}{A\arabic{figure}} % Figure counter representation
\renewcommand{\theHfigure}{A\arabic{figure}} % Hyperref figure hyperlink hook
\renewcommand{\thetable}{A\arabic{table}}
\renewcommand{\theHtable}{A\arabic{table}}
\makeatother
\renewcommand{\thetable}{A\arabic{table}}
\setcounter{mylemma}{0}
\renewcommand{\themylemma}{A\arabic{mylemma}}
\renewcommand{\thealgorithm}{A\arabic{algorithm}}
\setcounter{equation}{0}
\renewcommand{\theequation}{A\arabic{equation}}
% \counterwithin{table}{section}
% \counterwithin{figure}{section}

\section{Safety Under Different Compromised-Agent Placements}
\label{app:motivation}

\subsection{Configuration}
\label{app:motiv_setup}
Generation uses temperature $0.6$, $\mathrm{top}\text{-}p\,0.95$, and a budget of $4096$ tokens per agent. 
Fig.\,\ref{fig:motivation}(b) reports the attack success rate on HarmBench prompts \citep{mazeika2024harmbench}, StrongReject prompts \citep{souly2024strongreject}, and JailbreakBench prompts \citep{chao2024jailbreakbench}. 
These benchmarks do not share a criterion: 
Following the popular setttings, HarmBench is scored by its Llama-2-13B classifier, which asks whether the final answer is \emph{harmful}, whereas StrongReject and JailbreakBench are scored by whether the system \emph{complied} with the request rather than refusing it.

\subsection{Where the Compromised Agent Sits}
\label{app:seats}
\begin{table}[h]
\centering
\caption{\small{\textbf{Safety of an untrained MAS as a function of which agent is compromised.} No training anywhere: the two architectures of Fig.\,\ref{fig:motivation} run the safety-aligned backbone at every seat except the listed ones, which run the unaligned copy with the same role prompt. Attack success rate reported on all $159$ HarmBench prompts under the same judge as Table\,\ref{tab:safety_main}. The {\setlength{\fboxsep}{1pt}\colorbox{gray!15}{shaded}} rows are the compromised settings carried into Table\,\ref{tab:safety_main}.}}
\label{tab:seats}
\small
\setlength{\tabcolsep}{6pt}
\begin{tabular}{@{}llc@{}}
\toprule[1pt]
System & Compromised agent(s) & HarmBench ASR (\%\,$\downarrow$) \\
\midrule
\multirow{2}{*}{Single LLM} & none & 16.98 \\
 & the model itself & 72.96 \\
\midrule
\multirow{9}{*}{\textit{Sequential}} & none (benign MAS) & 17.61 \\
\cmidrule(l){2-3}
 & \cellcolor{gray!15}Solver & \cellcolor{gray!15}42.77 \\
 & Verifier & 19.50 \\
 & Corrector & 20.75 \\
 & Decider & 18.87 \\
\cmidrule(l){2-3}
 & Verifier $+$ Corrector & 22.01 \\
 & Verifier $+$ Decider & 25.16 \\
 & Verifier $+$ Corrector $+$ Decider & 40.88 \\
 & all four & 54.72 \\
\midrule
\multirow{6}{*}{\textit{Hierarchical}} & none (benign MAS) & 19.50 \\
\cmidrule(l){2-3}
 & Analyzer 1 & 27.04 \\
 & Analyzer 2 & 22.64 \\
 & Analyzer 3 & 25.79 \\
 & \cellcolor{gray!15}Synthesizer & \cellcolor{gray!15}52.83 \\
\cmidrule(l){2-3}
 & all four & 72.96 \\
\bottomrule[1pt]
\end{tabular}
\end{table}

Fig.\,\ref{fig:motivation} places the compromised agent at the position that degrades each architecture most: the Solver in the sequential chain and the Synthesizer in the hierarchy. \textbf{Table~\ref{tab:seats}} reports every other position, and every combination we ran, on the same untrained systems.
Fig.~2 uses the Solver in the sequential chain and the Synthesizer in the hierarchy as the compromised positions carried into the main experiments.
Table~\ref{tab:seats} reports the other compromised positions on the same untrained systems.
In the sequential MAS, compromising the Solver yields the largest observed ASR among the single-position settings ($42.77\%$ versus $17.61\%$ for the benign MAS), whereas compromising the Verifier, Corrector, or Decider produces smaller changes in this run.
In the hierarchical MAS, the largest observed ASR occurs when the Synthesizer is compromised ($52.83\%$ versus $19.50\%$ for the benign MAS), while compromising individual Analyzers results in more modest increases.
We do not interpret the small differences among the remaining positions as a stable ordering, since each configuration is evaluated with a single run.
Similarly, compromising the three downstream agents in the sequential MAS gives $40.88\%$ ASR, close to the $42.77\%$ observed with the Solver alone, so we avoid drawing conclusions from their ordering.
These results motivate training all roles rather than assuming in advance which position will be compromised.

\section{Details of the ADMM Algorithm}
\label{app:algo}

Algorithm~\ref{alg:admm} states one outer iteration of the procedure of Sec.~\ref{sec:admm}. The three steps follow that derivation: the $\mathbf x$-step is role-local and runs in parallel, the $\mathbf z$-step exchanges optimization variables across nodes, and the dual step is local.

\begin{algorithm}[htb]
\caption{Distributed MAS training via consensus ADMM}
\label{alg:admm}
\begin{algorithmic}[1]
\Require agents $i=1,\dots,N$ with overlaps $\Omega_i$, penalty $\rho>0$, local steps $L$, rounds $T$
\State initialize $\mathbf{x}_i^{(0)}=\mathbf{z}_i^{(0)}$ and $\mathbf{u}_i^{(0)}=\mathbf{0}$ for every agent $i$
\For{$t=0,1,\dots,T-1$}
  \State deploy $\{\mathbf{z}_i^{(t)}\}$ to the MAS, roll it out on queries $q\sim\mathcal{D}$, and score the rollouts for $r_i$
 %  \State \textbf{$\mathbf x$-step (role-local fine-tuning, in parallel)}: each agent $i$ takes $L$ gradient steps on \eqref{eq:mas_local_update} with
 % \hspace*{5mm} $f_i$ given by \eqref{eq:local_FT}, over its own $\btheta_i$ only, giving $\mathbf{x}_i^{(t+1)}$
\State \textbf{$\mathbf{x}$-step (role-local fine-tuning, in parallel):} each agent $i$ takes $L$ gradient steps on \eqref{eq:mas_local_update} 
\hspace*{5mm}with $f_i$ given by \eqref{eq:local_FT}, updating its local adapter copy $\btheta_i$ and obtaining $\mathbf{x}_i^{(t+1)}$
  \State \textbf{$\mathbf z$-step (consensus averaging)}: agents exchange $\mathbf b_i=\mathbf{x}^{(t+1)}_i+\mathbf{u}^{(t)}_i$ and form $\mathbf{z}_i^{(t+1)}$  \hspace*{5mm} blockwise by \eqref{eq:block_consensus_average}, using all-reduce or gossip protocol %of \textbf{Fig.}~\ref{fig:protocols}
  \State \textbf{dual step (in parallel)}: $\mathbf{u}_i^{(t+1)}\leftarrow\mathbf{u}_i^{(t)}+\mathbf{x}_i^{(t+1)}-\mathbf{z}_i^{(t+1)}$ for every agent $i$
\EndFor
\State \Return the consensus variables $\{\mathbf{z}_i^{(T)}\}$ %, which are what the MAS deploys
\end{algorithmic}
\end{algorithm}

\subsection{Initialization of $\mathbf x$, $\mathbf z$ and $\mathbf u$}
\label{app:init}

Algorithm~\ref{alg:admm} opens with
$\mathbf x_i^{(0)}=\mathbf z_i^{(0)}$ and
$\mathbf u_i^{(0)}=\mathbf 0$.
These variables are initialized once before the first ADMM round and are then carried across subsequent rounds.

\textbf{The dual $\mathbf u_i$ starts at zero.}
We use the scaled dual, $\mathbf u_i=\mathbf y_i/\rho$, where $\mathbf y_i$ is the Lagrange multiplier associated with the constraint $\mathbf x_i=\mathbf z_i$.
Before the first round, there is no accumulated disagreement to carry, so we initialize $\mathbf u_i^{(0)}=\mathbf 0$.
The ablation that removes the dual (Sec.~\ref{sec:exp}) keeps this variable fixed at zero throughout the run.

\textbf{The primal variables start at the deployed adapter, which is zero only for an untrained system.}
The optimization variable $\btheta_i$ is the trainable factor of the LoRA adapter, and both $\mathbf x_i$ and $\mathbf z_i$ are initialized from the adapter currently deployed by the system.
Thus, $\mathbf x_i^{(0)}=\mathbf z_i^{(0)}$ holds by construction, and the initial primal residual is exactly zero.
Because we train only the up-projection $B_i$ and $B_i$ is zero-initialized in the standard LoRA scheme, a run starting from the untrained system has
$\mathbf x_i^{(0)}=\mathbf z_i^{(0)}=\mathbf 0$:
the deployed update $\Delta W_i=B_iA$ vanishes, and round $0$ corresponds to the backbone itself.
A run warm-started from a checkpoint instead begins from its (possibly nonzero) $B_i$.

\textbf{The frozen factor $A$ is not an optimization variable.}
$A$ is initialized once, shared across agents, and never updated, so it does not enter $\mathbf x$, $\mathbf z$, or $\mathbf u$.
This also makes the $\mathbf z$-step exact in weight space (App.~\ref{app:lora}), since averaging the trainable factors $B_i$ is equivalent to averaging the corresponding updates $\Delta W_i=B_iA$.
It is also why zeroing both LoRA factors would be inappropriate in a variant that trains both factors: if both $A$ and $B$ were initialized to zero, then $\Delta W=BA$ and the gradients with respect to both factors would initially vanish.

\textbf{Each $\mathbf{x}$-step starts from the current proximal anchor.}
At ADMM round $t$, the local $\mathbf{x}$-step is initialized at
$\mathbf a_i=\mathbf z_i^{(t)}-\mathbf u_i^{(t)}$
rather than at $\mathbf x_i^{(t)}$.
Agent $i$ then performs gradient-based optimization of its local adapter copy $\btheta_i$ using the role-local objective, with the sampled rollout and received messages treated as fixed training context.
Starting from $\mathbf a_i$ makes the quadratic penalty zero at the first inner step and nonzero only as the solution moves away from the anchor, preventing the local update from being dominated by the absolute scale of $\rho$.
The resulting $\mathbf{x}_i^{(t+1)}$, together with the subsequent $\mathbf z_i^{(t+1)}$ and $\mathbf u_i^{(t+1)}$, is carried to the next ADMM round.

\section{Experimental Details}
\label{app:exp_details}
% This appendix collects the setup details deferred from Sec.~\ref{sec:setup}.

\subsection{System Prompts}
\label{app:prompts}
% Every agent receives the same base system message and the same answer-format instruction, byte for byte, as the single-agent baseline; a comparison between arms that were told to format differently would not be meaningful. The roles differ only in the instruction that follows.

In the sequential chain, the \textbf{Solver} is asked to solve the question completely, to number its steps and state what each step establishes so that any single step can be checked in isolation, and to commit to a specific answer rather than hedge. The \textbf{Verifier} is shown the candidate solution and told that its only job is to find errors, checking in order whether the candidate answers the question actually asked, whether the method is the right one, whether each algebraic step is correct, whether units, signs and rounding are right, and whether the answer is in the demanded form. The \textbf{Corrector} receives the candidate together with the verifier report and repairs the identified errors. The \textbf{Decider} sees the prior agents' work and issues the final answer.

In the hierarchy, the three \textbf{Analyzers} are told that they work on the same question independently and cannot see one another, so that their value lies in reaching the answer by different routes: the first derives from principles before looking at the options, the second proceeds by elimination, and the third tests candidate answers on concrete values. The \textbf{Synthesizer} compares all three analyses and decides the final answer.
The verbatim prompts for both architectures are released with our code.

\subsection{Backbone and Adapter}
\label{app:lora}

The aligned backbone is \texttt{Qwen/Qwen3-8B}, run in thinking mode with a fixed generation budget.
The compromised agent uses \texttt{mlabonne/Qwen3-8B-abliterated}, an abliterated variant with reduced refusal behavior~\citep{arditi2024refusal}.
It shares the tokenizer and architecture of the aligned backbone and can therefore replace any role without architectural changes.
Each agent maintains a local trainable parameter copy $\btheta_i$, implemented as a rank-$16$ LoRA adapter with scaling $\alpha=16$, applied to the seven projections (\texttt{q}, \texttt{k}, \texttt{v}, \texttt{o}, \texttt{gate}, \texttt{up}, and \texttt{down}) in all $36$ layers.
Writing the corresponding local update to a base weight as
$\Delta W_i=B_iA$, we initialize the down-projection $A$ identically across agents and keep it frozen, while training only $B_i$.

\subsection{Details of training}
\label{app:collect}

\paragraph{Training data.}
\textbf{Harmful requests.}
We construct the harmful training set from AdvBench~\citep{zou2023advbench} and the standard and contextual HarmBench validation splits~\citep{mazeika2024harmbench}, assigning all retained prompts a refusal target.
We remove prompts overlapping with HarmBench or XSTest evaluation data, and use only HarmBench validation examples that are disjoint from the test splits used for evaluation.

\textbf{Benign requests.} A slice of the MMLU-Pro test split~\citep{wang2024mmlupro}, kept with their native QA answer type and scored as ordinary questions. They are the over-refusal / stay-helpful control: without them every retained fine-tuning target is a refusal and the trained system collapses into refusing everything. 

\textbf{Safety.} 
% Collection runs once over the assembled pool. The scaffold used at collection time carries a safety-hardened variant of the role prompts, and the compromised agent is present throughout: its backbone is the abliterated copy, it follows the protocol, and it loads the deployed adapter like every other agent, so the recorded trajectories show the system as it actually behaves under compromise. 
Each agent is then fit on its message with the \emph{ordinary} deployment prompt as input, so what is learned is the behaviour itself rather than the collection prompt, and nothing at test time depends on the hardened prompt. We draw up to eight trajectories per query, at temperature $0.8$ and $\mathrm{top}\text{-}p\,0.95$, with a per-agent budget of $4096$ tokens.

\textbf{General reasoning.} Collection is on-policy and repeats every outer round. 
% Before each round the current consensus variable is loaded into the system, so the new rollouts come from the policy being trained rather than from the frozen backbone. 
Queries are selected by excluding every anchor already used in an earlier round, so successive rounds never overlap; each round pre-screens a window of candidates by sampling each twice and dropping those answered correctly both times, then collects a handful of trajectories for each survivor and keeps only the groups whose outcomes are mixed. 
Sampling uses the same temperature and $\mathrm{top}\text{-}p$ with an $8192$-token per-agent budget. 
% Note that a round therefore changes both the policy and the query set; an ablation that isolates the policy effect would hold the queries fixed.

\paragraph{Optimization.}
Table~\ref{tab:hparams} lists the optimizer settings.
% ; centralized and distributed runs share every value except the optimizer itself, so that a paradigm difference is never a hyperparameter difference. 
Each ADMM round deploys the current consensus adapter, collects (or reuses) a batch of rollouts, and runs $L$ local gradient steps in the $\mathbf x$-step of \eqref{eq:mas_local_update}.

\begin{table}[t]
\centering
\caption{\small{Optimizer and ADMM settings, read off the trained checkpoints. Distributed and centralized runs share every value; the penalty $\rho$ and the round count apply to the distributed runs. 
% Both lines use a preference objective.
}}
\label{tab:hparams}
\small
\begin{tabular}{@{}lcc@{}}
\toprule
 & Safety & Reasoning \\
\midrule
ADMM rounds $T$                       & $24$ (H-MAS) / $20$ (S-MAS) & $8$ \\
Local gradient steps $L$ per $\mathbf x$-step & $16$      & $16$       \\
Rollouts per query                    & $8$        & $8$        \\
Optimizer                             & Adam       & Adam       \\
Learning rate                         & $10^{-3}$  & $10^{-3}$  \\
Gradient clip                         & $0.01$     & $0.01$     \\
Batch size                            & $8$        & $8$        \\
ADMM penalty $\rho$                   & $0.03$     & $0.1$      \\
DPO temperature $\beta_{\mathrm{DPO}}$ & $0.1$     & $0.1$        \\
KL weight $\beta_{\mathrm{KL}}$       & $0.1$      & $0.1$      \\
Role-local loss                       & DPO        & DPO \\
\bottomrule
\end{tabular}
\end{table}

% \paragraph{Data filtering.}
\paragraph{Data filtering.}
Each role is trained using the system-level outcome reward of the full rollout~\citep{wan2025rema,marft2025,liu2026magrpo}.
For reasoning, $R$ is answer correctness, and we retain queries whose sampled rollouts contain both successful and unsuccessful outcomes.
For safety, harmful queries receive $R=1$ when the final response is judged harmless, while benign queries receive $R=1$ when answered correctly; refusals on benign queries are therefore penalized.
We further include a role's message in the safety loss only when the rollout succeeds and the message itself is harmless for harmful queries or non-refusal for benign queries, preventing unsafe intermediate messages from being reinforced by a safe final response.
All roles remain trainable, including the compromised role.
For both tasks, we balance training examples across roles, de-duplicate repeated targets, and discard generations with incomplete reasoning blocks.
By default, the loss is applied to the full message, including its reasoning content.

\subsection{Details of evaluation}
\label{app:eval}

The five safety benchmarks use different scoring criteria, so their absolute values are not directly comparable.
HarmBench uses its Llama-2-13B classifier~\citep{mazeika2024harmbench} to judge whether the final response is \emph{harmful}, whereas StrongReject, JailbreakBench, and toxic OR-Bench measure whether the system \emph{complies} rather than refuses.

The HarmBench Llama-2-13B classifier is the primary harmfulness judge throughout, pinned to one revision. A Granite guardian classifier was run as a sensitivity check and is not used for any reported number: it flags terse refusals as harmful and shifts the comparison by roughly five points, in a direction that depends on how verbose an arm is. Where a benchmark ships its own notion of compliance we keep it, so that our numbers stay comparable to how that benchmark is normally reported, and we never compare an absolute value across benchmarks.

Evaluation runs in thinking mode at temperature $0.6$ and $\mathrm{top}\text{-}p\,0.95$ with a fixed seed. 
Two fixed profiles are used and never mixed: safety evaluations allow $4096$ tokens per agent and capability evaluations allow $8192$. 

% All comparisons are paired row by row on the same items. A pair is admitted only when the two arms agree on architecture, agent count, compromised position, evaluation profile and hardware; mismatches on any of these have produced apparent effects larger than the effects of interest. Significance is reported from paired tests over the per-row outcomes, and the untrained reference for each architecture is pinned once and reused rather than re-selected per arm.
For general reasoning, we evaluate each benchmark on a fixed subset sampled once and reused across all methods, ensuring that every bar in Fig.~\ref{fig:general_bars} uses the same questions.
Each subset contains up to $256$ examples.
Safety benchmarks use their standard evaluation splits.

\section{Scaling with the Number of Agents and the Backbone}
\label{app:scaling}

We further examine the compromised-agent effect across different system sizes and backbone sizes.
These experiments are intended as supplementary analyses beyond the main $N=4$, \texttt{Qwen3-8B} setting in Table~\ref{tab:safety_main}.

\paragraph{Scaling with the number of agents.}
Table~\ref{tab:nsweep} evaluates untrained MAS with different numbers of agents.
For S-MAS, we consider $N\in\{2,3,5,6,7\}$ with the Solver compromised; for H-MAS, we consider $N\in\{3,5,6,8\}$ with the Synthesizer compromised.
The main $N=4$ configuration is reported separately in Table~\ref{tab:safety_main}.
Across all tested configurations, introducing one compromised agent increases HarmBench ASR relative to the corresponding benign system, with observed increases ranging from $14.47$ to $32.08$ percentage points.
The magnitude of this increase varies across system sizes without a clear monotonic trend in $N$.

\begin{table}[h]
\centering
\caption{\small{
\textbf{Effect of one compromised agent across system sizes.}
We evaluate untrained S-MAS with $N\in\{2,3,5,6,7\}$ and H-MAS with $N\in\{3,5,6,8\}$.
The compromised agent is the Solver in S-MAS and the Synthesizer in H-MAS.
The main $N=4$ configuration is reported separately in Table~\ref{tab:safety_main}.
We report HarmBench ASR (\%) and the compromised--benign difference $\Delta$.
}}
\label{tab:nsweep}
\small
\setlength{\tabcolsep}{6pt}
\begin{tabular}{@{}lcccc@{}}
\toprule[1pt]
System & $N$ & Benign & Compromised & $\Delta$ \\
\midrule
\multirow{5}{*}{\textit{S-MAS}}
& $2$ & 15.09 & 29.56 & $+14.47$ \\
& $3$ & 20.13 & 40.25 & $+20.13$ \\
& $5$ & 16.98 & 38.36 & $+21.38$ \\
& $6$ & 22.01 & 40.88 & $+18.87$ \\
& $7$ & 16.98 & 49.06 & $+32.08$ \\
\midrule
\multirow{4}{*}{\textit{H-MAS}}
& $3$ & 18.87 & 45.28 & $+26.42$ \\
& $5$ & 25.16 & 47.80 & $+22.64$ \\
& $6$ & 23.27 & 47.17 & $+23.90$ \\
& $8$ & 23.27 & 51.57 & $+28.30$ \\
\bottomrule[1pt]
\end{tabular}
\end{table}

\paragraph{Training beyond the main system size.}
We additionally evaluate compromised S-MAS at $N=3$, whose corresponding untrained configuration is included in Table~\ref{tab:nsweep}.
As shown in Table~\ref{tab:nsweep_train}, centralized and distributed training reduce HarmBench ASR by $5.03$ and $3.14$ percentage points, respectively, relative to the untrained system.
We report this result as an additional evaluation beyond the main $N=4$ setting rather than drawing conclusions about how training effectiveness scales with the number of agents.

\begin{table}[h]
\centering
\caption{\small{
\textbf{Additional training result beyond the main system size.}
Change in HarmBench ASR (percentage points; negative is safer) relative to the corresponding untrained compromised S-MAS in Table~\ref{tab:nsweep}.
}}
\label{tab:nsweep_train}
\small
\setlength{\tabcolsep}{9pt}
\begin{tabular}{@{}lcc@{}}
\toprule[1pt]
$N$ & Centralized $\Delta$ ASR & Distributed $\Delta$ ASR \\
\midrule
$3$ & $-5.03$ & $-3.14$ \\
\bottomrule[1pt]
\end{tabular}
\end{table}

\paragraph{Effect on a smaller backbone.}
We further evaluate an untrained $N=4$ S-MAS using \texttt{Qwen3-4B}.
As shown in Table~\ref{tab:backbone}, compromising the Solver increases HarmBench ASR from $13.84\%$ to $48.43\%$, corresponding to a $34.59$-point increase.
The corresponding \texttt{Qwen3-8B} results are reported in the main experiment in Table~\ref{tab:safety_main}.
Together, these experiments show that the compromised-agent effect is also present with a smaller backbone.

\begin{table}[h]
\centering
\caption{\small{
\textbf{Effect of one compromised agent on a smaller backbone.}
We evaluate an untrained S-MAS with $N=4$ using \texttt{Qwen3-4B}, with the Solver as the compromised agent.
We report HarmBench ASR (\%) and the compromised--benign difference $\Delta$.
The corresponding \texttt{Qwen3-8B} results are reported in Table~\ref{tab:safety_main}.
}}
\label{tab:backbone}
\small
\setlength{\tabcolsep}{9pt}
\begin{tabular}{@{}lccc@{}}
\toprule[1pt]
Backbone & Benign & Compromised & $\Delta$ \\
\midrule
\texttt{Qwen3-4B} & 13.84 & 48.43 & $+34.59$ \\
\bottomrule[1pt]
\end{tabular}
\end{table}

\end{document}